\ifx\pdfoutput\undefined\else\pdfoutput=1\fi
\documentclass[11pt]{article}

\usepackage[preprint]{acl} % Keeps the original ACL style sheet active

\usepackage[LAE,T1]{fontenc}       % Enables Arabic (LAE) and English (T1) encodings
\usepackage[utf8]{inputenc}        % EXPLICITLY REQUIRED for typing raw Arabic characters under pdfLaTeX
\usepackage[arabic,english]{babel} % Sets English as main language, Arabic as secondary

\usepackage{latexsym}
\usepackage{amsmath, amssymb, amsthm}
\usepackage{graphicx}
\usepackage{booktabs}
\usepackage{microtype}
\usepackage{enumitem}
\usepackage{url}
\usepackage{xcolor}
\usepackage{hyperref}

\newcommand{\zlex}{\ensuremath{z_{\text{lex}}}}
\newcommand{\zcult}{\ensuremath{z_{\text{cult}}}}
\newcommand{\zmet}{\ensuremath{z_{\text{met}}}}
\newcommand{\Wmet}{\ensuremath{W_{\text{met}}}}
\newcommand{\Wlex}{\ensuremath{W_{\text{lex}}}}
\newcommand{\Wcult}{\ensuremath{W_{\text{cult}}}}
\newcommand{\Rhier}{\ensuremath{R_{\text{hier}}}}

\newcommand{\arb}[1]{\textAR{#1}}

\title{CAMMAR: Culture-Aware Matryoshka for Metaphorical Arabic Representations}

\author{%
  Suzan Awinat\thanks{Corresponding author.} \\
  \small
  Department of Computer Engineering \\
  Autonomous University of Madrid (UAM) \\
  School of Science and Technology \\
  IE University \\
  Madrid, Spain \\
  \texttt{suzan.tayseer@estudiante.uam.es}
  \And
  Alfonso Ortega de la Puente \\
  Department of Computer Science \\
  Oviedo University \\
  Asturias, Spain
}

\begin{document}

\maketitle

% --- Abstract -------------------------------------------------------------
\begin{abstract}
Metaphor in Arabic is a culturally grounded mechanism for constructing meaning, encoding cultural knowledge that shapes interpretation. Yet current Arabic language models typically collapse lexical, cultural, and metaphorical information into a single representational space, a phenomenon we term "semantic smearing". We introduce CAMMAR (Culture-Aware Matryoshka for Metaphorical Arabic Representations), a representation learning framework that organizes meaning into nested lexical, cultural, and metaphorical embedding subspaces through a staged semantic curriculum. The design implements compositional principles of Al-Jurj\=an\={\i}'s theory of \emph{na\d{z}m} (\arb{نظم}), modeling figurative meaning as compositionally grounded in prior semantic relations, and yields a training-free geometric measure of metaphoricity based on the distance between lexical and metaphorical representations.

Evaluated on a new span-annotated Arabic metaphor set as word-matched figurative/literal pairs, the geometric readout detects metaphor well above chance when the inter-layer geometry is shaped by paired supervision (AUC up to 0.84; figurative outscores its literal counterpart for the same word in 82.6\% of pairs), but sits at chance under an unsupervised domain contrast alone, a clean separation between a legible-under-supervision regime and a non-emergent one. A controlled ablation shows that grounding the lexical layer in morphological roots gives a small but consistent gain, an effect absent from direct probing that reflects the layer's quality as a measurement anchor. We will release the datasets, cultural concept inventory, and code upon acceptance.
\end{abstract}

% ============================================================================
\section{Introduction}
\label{sec:intro}
% ============================================================================

Metaphor is one of the most persistent challenges in Natural Language Processing. Despite the recent progress of large language models (LLMs) on most generative and discriminative benchmarks, multiple evaluations have shown that figurative-language understanding remains substantially below human level, particularly under culturally dense or rhetorically complex contexts \citep{ZHANG2025, sanchezbayona2025metaphors, mangiaterra2026metaphors}. The problem is especially acute for Arabic. Recent benchmarks designed specifically for Arabic figurative language report systematic failures across both general-purpose and Arabic-specialized models. \emph{Fann or Flop} \citep{alghallabi2025fann} shows that state-of-the-art LLMs cannot reliably capture the interpretive depth of Classical Arabic poetry across historical eras. \citet{attia2026pragmatic} document a drop of more than 14 points in accuracy when 22 LLMs are evaluated on the \emph{pragmatic use} of culturally grounded Arabic idioms compared to multiple-choice understanding of the same expressions. These findings suggest that the gap is not a matter of scale, but of how metaphorical meaning is represented internally.

We hypothesize that one important structural cause is the following. Standard dense embeddings compress lexical meaning, cultural symbolism, and metaphorical abstraction into a single shared space a phenomenon we call \emph{semantic smearing}. This is acute for Arabic, where one triconsonantal root generates both concrete lexical items and conventionalized metaphorical extensions: \arb{أ-س-د} gives both \emph{asad} (lion, the animal) and its figurative sense of bravery and leadership.

The Arabic rhetorical tradition (\emph{al-Bal\=agha}, \arb{البلاغة}) has distinguished literal (\emph{\d{h}aq\={\i}qa}) from metaphorical (\emph{maj\=az}) meaning for over a millennium. Its foundational figure, \citet{jurjanidalail}, argues that meaning resides not in individual lexical items but in their relations: his concept of \emph{na\d{z}m} (\arb{نظم}) holds that meaning emerges from structured composition, and his treatment of metaphor \citep{jurjaniasrar} grounds figurative interpretation in a relation (\emph{'al\=aqa}) between the figurative reading and its literal anchor, a view elaborated by \citet{zamakhshariasas} and \citet{sakkakimiftah} and paralleled by modern accounts of metaphor as cross-domain projection \citep{lakoff1980metaphors}. We operationalize this compositional, relational view within a representation-learning framework.

We propose CAMMAR (Culture-Aware Matryoshka for Metaphorical Arabic Representations), which organizes Arabic figurative meaning into three nested embedding subspaces of decreasing dimensionality: outer lexical, middle cultural, inner metaphorical, sharing one Arabic backbone and trained by a three-phase curriculum that protects earlier structure while building later abstraction. Unlike standard Matryoshka Representation Learning \citep{kusupati2024matryoshka}, which supervises all prefixes identically, CAMMAR assigns progressively abstract objectives to progressively inner prefixes. From this structure we derive a continuous \emph{metaphorical weight} \Wmet{}, read out at inference as a geometric divergence between the outer and inner layers, giving an interpretable per-word metaphoricity estimate with no supervised classifier.

We evaluate CAMMAR on an author-verified span-annotated Arabic metaphor test set. Our experiments examine whether metaphoricity is legible in the inter-layer geometry and under what supervision, and whether grounding the lexical layer in morphological roots improves the readout.

Our contributions are:
\begin{itemize}[leftmargin=*, itemsep=0pt]
  \item \textbf{CAMMAR}, a sequential nested supervision framework that imposes a curriculum of progressively abstract semantic objectives along the nested prefixes of a single Arabic encoder.
  \item A formalization of \emph{semantic smearing} in figurative-language embeddings, together with a hierarchical regularization strategy that mitigates it through controlled freezing across training phases.
  \item A geometric \emph{metaphorical weight} \Wmet{} that estimates per-word metaphoricity through prefix divergence, without requiring example-level metaphor annotations at training time.
  \item A small but carefully constructed author-verified evaluation set for Arabic metaphor at the target-word level, to be released alongside the model.
\end{itemize}

% ============================================================================
\section{Related Work}
\label{sec:related}
% ============================================================================

CAMMAR sits at the intersection of computational metaphor processing, Arabic cultural-semantic modeling, and nested representation learning.

\paragraph{Computational metaphor processing.} Neural metaphor processing \citep{shutova2010survey} has moved from feature engineering toward contextual representations \citep{dodinh2016token, choi2021melbert}, with recent work treating metaphoricity as a context-sensitive discrepancy between a word's literal meaning and its context \citep{he2024scnet, jia2024metaphor}. CAMMAR shares this intuition but internalizes it \emph{within} a single nested embedding rather than across two models. The framing is motivated by evaluations showing scale alone does not solve metaphor: LLMs rely on surface heuristics \citep{sanchezbayona2025metaphors}, achieve low detection F1 \citep{boisson2024metaphor}, and misjudge novel metaphors \citep{mangiaterra2026metaphors}.

\paragraph{Arabic figurative language and the cultural gap.} In Arabic, metaphor is entangled with morphology and inherited symbolism. Benchmarks show state-of-the-art models failing on Classical Arabic poetry \citep{alghallabi2025fann} and on culturally grounded idioms and proverbs \citep{magdy2025jawaher, attia2026pragmatic, zibin2025arabic}. Prior Arabic metaphor work is largely supervised binary classification on small datasets, or symbolic resources \citep{banou2025multilayer} that treat figuration as a downstream application. To our knowledge no prior work models Arabic figurative meaning through a \emph{hierarchical learned representation} separating lexical, cultural, and figurative content within one geometry. Cross-lingual studies reinforce that figurative meaning depends on language-specific cultural grounding rather than surface transfer \citep{sanchezbayona2026meta4xnli, tourajmehr2025persian}.

\paragraph{Nested representation learning.} Matryoshka Representation Learning \citep{kusupati2024matryoshka} nests embeddings under a single identical objective, encouraging compactness, not specialization; \citet{lai2026mrl4rec} show identical-supervision prefixes differ only in gradient magnitude, giving no directional differentiation. Extensions add sequential training and adaptive dimension selection \citep{zhang2025smec} and strong Arabic nested embeddings \citep{nacar2024enhancing}, but their curriculum is over \emph{compression ratios}; CAMMAR's is over \emph{semantic abstractions}, assigning different objectives to different prefixes. Closest in spirit, CAMEL \citep{zhang2024camel} disentangles literal and metaphorical meaning \emph{across} two encoders with cross-domain attention, whereas CAMMAR organizes it \emph{within} one nested embedding; we adopt CAMEL's self-paced contrastive temperature schedule in Phase~3.

% ============================================================================
\section{Methodology}
\label{sec:method}
% ============================================================================

\subsection{Overview}
\label{sec:overview}

\begin{figure*}[t]
  \centering
  \includegraphics[width=0.88\textwidth]{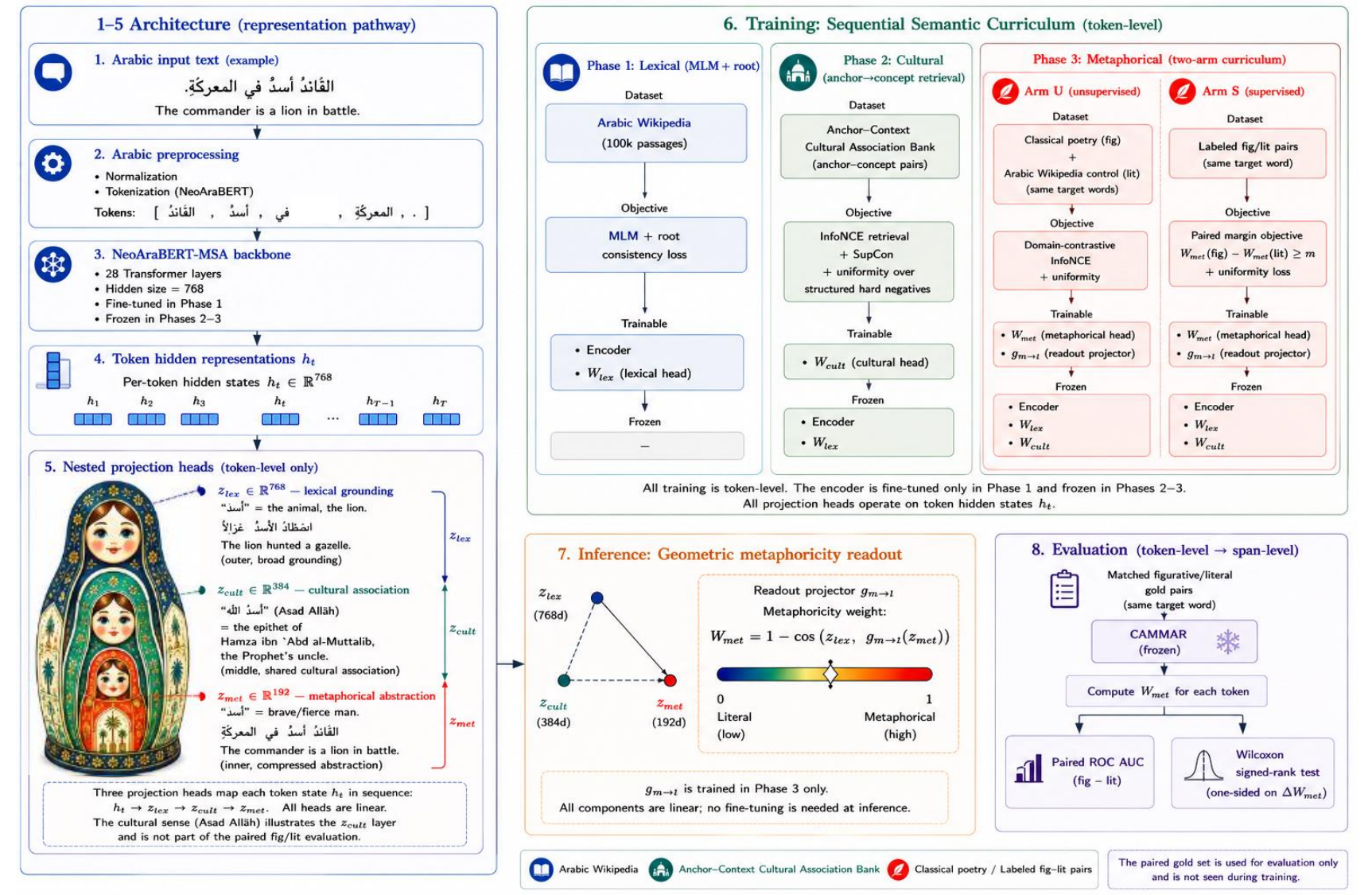}
  \caption{End-to-end CAMMAR pipeline: paired figurative/literal inputs are encoded by a shared NeoAraBERT backbone, projected into nested lexical (\zlex{}), cultural (\zcult{}), and metaphorical (\zmet{}) subspaces, trained by a three-phase curriculum with controlled freezing, and read out geometrically for span-level evaluation on matched pairs. The architecture is entirely token-level; target-word representations aggregate token embeddings over the annotated span.}
  \label{fig:pipeline}
\end{figure*}

We propose CAMMAR, which organizes figurative meaning into three nested embedding subspaces of decreasing dimensionality, outer \zlex{} biased toward lexical surface form, middle \zcult{} toward culturally conditioned meaning, inner \zmet{} toward figurative abstraction, sharing parameters through a single Arabic backbone and trained by a three-phase curriculum that protects earlier structure while building later abstraction (Fig.~\ref{fig:pipeline}). This operationalizes \citeposs{jurjanidalail} \emph{na\d{z}m}: the metaphorical layer acquires content only through its dependence on the cultural and lexical layers that constitute it, and the controlled freezing across phases reflects al-Jurj\=an\={\i}'s insistence that figurative interpretation remains grounded in, rather than displacing, prior semantic relations. The layers nest generatively: \zmet{} is computed from \zcult{}, which is computed from \zlex{}, so figurative interpretation is routed through cultural meaning (\arb{سراب} as optical mirage $\to$ deceptive promise $\to$ false hope) rather than read directly off lexical form. We add one non-Jurj\=an\={\i}an element: an explicit cultural-substrate layer modeling the shared knowledge that Classical Arabic interpretation presupposes.

A key property is that the projection heads operate at the \emph{token level},  producing one nested embedding per token, which is what enables per-target-word metaphoricity estimation against a span-annotated gold set. The same heads can also be applied to a pooled representation for sentence-level analysis, but all training objectives and the evaluation in this work are token-level. We describe the architecture in \S\ref{sec:arch}, the curriculum and objectives in \S\ref{sec:curriculum}-\S\ref{sec:phases}, the metaphor weight in \S\ref{sec:wmet}, and the evaluation protocol in \S\ref{sec:eval}.

\subsection{Architecture}
\label{sec:arch}

\paragraph{Backbone.}
We use NeoAraBERT-MSA \citep{abou-chakra-etal-2026-neoarabert} as the shared encoder, a 28-layer transformer ($d{=}768$) with a 65k SentencePiece vocabulary, pretrained on Modern Standard Arabic with robust coverage of both contemporary and Classical Arabic important for our metaphor corpus, which spans both registers.

We denote the encoder output for an input sequence $x = (x_1, \dots, x_L)$ as token-level hidden states $h_t \in \mathbb{R}^d$, $t = 1, \dots, L$.

\paragraph{Nested projection heads.}
Each token state is projected into three nested semantic subspaces of decreasing dimensionality:
\begin{align}
  \zlex(t)  &= \Wlex (h_t) \in \mathbb{R}^{768}, \\
  \zcult(t) &= \Wcult(\zlex(t)) \in \mathbb{R}^{384}, \\
  \zmet(t)  &= \Wmet (\zcult(t)) \in \mathbb{R}^{192}.
\end{align}
The composition $\zmet = \Wmet \circ \Wcult \circ \Wlex$ enforces a semantic nesting: information passes through the lexical and cultural layers before reaching the metaphorical one. Each head is linear, followed by GELU and layer normalization, kept shallow to preserve a tractable geometric interpretation. All objectives and evaluation operate on these token-level embeddings; the heads may also be applied to a pooled representation for sentence-level analysis. In our convention the outer (largest) layer is most concrete and inner layers progressively more abstraction through compression, the inverse of the dimension-as-specificity convention in some MRL work \citep[e.g.,][]{lai2026mrl4rec}.

Because the three layers live in different coordinate frames, the readout cannot compare \zlex{} and \zmet{} directly. We introduce a \emph{readout projector} $g_{\text{m}\to\text{l}}: \mathbb{R}^{192}\to\mathbb{R}^{768}$ (linear map then layer normalization) that maps \zmet{} back into the lexical frame, so metaphoricity is a within-frame divergence (\S\ref{sec:wmet}). It is trained \emph{only} in Phase~3 and excluded from the hierarchical regularizer; otherwise the readout would collapse to a constant.
\paragraph{Hierarchical re-projection.}
To regularize the geometry across layers, we add two learned re-projections $\pi_{\text{lex}}: \mathbb{R}^{384}\to\mathbb{R}^{768}$ and $\pi_{\text{cult}}: \mathbb{R}^{192}\to\mathbb{R}^{384}$, used by \Rhier{} (\S\ref{sec:phases}) and not at inference: a well-organized nested representation should allow approximate reconstruction of the outer layer from the inner one.

\subsection{Sequential Semantic Curriculum}
\label{sec:curriculum}

CAMMAR is trained in three phases of increasing abstraction (Fig.~\ref{fig:dataflow}): lexical grounding (Phase 1, training the encoder and \Wlex{}), cultural context (Phase 2, training \Wcult{} and $\pi_{\text{lex}}$ with earlier layers frozen), and figurative abstraction (Phase 3, training \Wmet{} and $\pi_{\text{cult}}$ with all else frozen). The encoder is fine-tuned only in Phase 1. This controlled freezing preserves earlier-phase structure while building more abstract representations on top; without it, a single end-to-end objective homogenizes the subspaces, as observed for standard MRL \citep{kusupati2024matryoshka, zhang2025smec, lai2026mrl4rec}.

\subsection{Phase-Specific Objectives}
\label{sec:phases}

\paragraph{Phase 1: lexical grounding.}
The first phase trains the encoder and \Wlex{} to capture lexical and morphological structure on Arabic Wikipedia (100k passages subsampled from $\sim$5.2M), combining two objectives.

The first is masked language modeling: we mask 15\% of input tokens following the standard BERT scheme \citep{devlin2019bert} and predict the masked token identity from the lexical projection via $\mathcal{L}_{\text{MLM}} = -\sum_{t\in\mathcal{M}} \log p(x_t \mid z_{\text{lex}}(t))$, tying the lexical layer to surface-form prediction.

The second is a root-consistency loss. Arabic is templatic: words sharing a triconsonantal root frequently share a core semantic field. We obtain roots from a morphological analyzer (ALMA/SINATools, with a CAMeL Tools fallback), computed offline and cached, rather than from a substring heuristic, since roots are not substrings of inflected forms. At the token level we apply a supervised-contrastive (InfoNCE-style) loss $\mathcal{L}_{\text{root}}$ over $z_{\text{lex}}$ that pulls together stem tokens sharing a root and pushes apart tokens with different roots (positives $P(i)$ = in-batch tokens sharing $i$'s root; alternate roots treated as equivalent). The total Phase~1 loss is $\mathcal{L}_1 = \mathcal{L}_{\text{MLM}} + \alpha \mathcal{L}_{\text{root}}$.

\paragraph{Phase 2: cultural association.}
The cultural layer must encode the conventional associations an Arabic speaker activates upon encountering a culturally loaded item (\arb{حاتم} $\mapsto$ generosity; \arb{عنترة} $\mapsto$ courage), since these associations are the ground (\emph{j\=ami\textquotesingle}) on which the figurative reading is built. We therefore train \Wcult{} not as a document-topic classifier but as an \emph{anchor-to-association} retriever (Appendix Fig.~\ref{fig:cultural}), on an Anchor Context Cultural Association Bank: short passages (1-3 sentences) in which an anchor entity is paired with the cultural concept its context activates, drawn from biographies, adab anthologies, proverb collections, poetry commentary, and heritage texts.\footnote{We deliberately learn cultural associations distributionally from text rather than from a symbolic knowledge graph; Future Work (\S\ref{sec:conclusion}) discusses a graph-based extension as future work.} Concepts are drawn from a fixed inventory $\mathcal{C}$ of 74 conventional cultural associations (e.g. \textsc{generosity}, \textsc{courage}, \textsc{treachery}, \textsc{patience}), including a \textsc{literal} class for non-activating contexts; the fixed inventory makes both same-concept negatives and concept-level evaluation well-posed.

Let $u = \operatorname{pool}_{t\in a}\, z_{\text{cult}}(t)$ be the anchor-span-pooled cultural vector and $p_k$ a learnable prototype for concept $k\in\mathcal{C}$. The primary objective is an InfoNCE anchor-to-concept retrieval $\mathcal{L}_{\text{a2c}}$ that matches $\hat u$ to its gold prototype $\hat p_{k^+}$ against all prototypes in $\mathcal{C}$, augmented by a supervised-contrastive term over \emph{structured} hard negatives the same anchor in an unrelated context, the same concept under a different anchor, and purely literal contexts and a uniformity penalty $\mathcal{L}_{\text{unif}}$ that prevents the projection from collapsing onto a low-rank cone. We apply the hierarchical regularizer $R_{\text{hier}}^{\text{cult}} = \lVert \pi_{\text{lex}}(\zcult) - \zlex \rVert_2^2$ as in Phase~1, and freeze the encoder and \Wlex{} so the literal layer is preserved. The full objective is $\mathcal{L}_2 = \mathcal{L}_{\text{a2c}} + \lambda_{\text{scl}}\mathcal{L}_{\text{scl}} + \lambda_{\text{unif}}\mathcal{L}_{\text{unif}} + \beta R_{\text{hier}}^{\text{cult}}$. Concept labels for training are LLM-proposed (distant supervision); reported cultural-layer results use a human-verified gold subset disjoint from training at the lemma and root level. We present the cultural layer as a method and resource; its empirical evaluation (held-out concept retrieval) is reported in \S\ref{sec:results}.

\paragraph{Phase 3: figurative abstraction.}
Phase 3 trains the metaphorical layer \Wmet{} and the readout projector $g_{\text{m}\to\text{l}}$ with the encoder, \Wlex{}, and \Wcult{} frozen, in one of two arms (\S\ref{sec:results-arms}). \textbf{Arm~U} (unsupervised) derives supervision from corpus identity, treating figurative-rich classical poetry ($\sim$330k passages) as one domain and a sampled Arabic Wikipedia literal control ($\sim$30\% as many) as another, and separating them in \zmet{} with a domain-contrastive InfoNCE objective \citep{oord2018cpc} and a uniformity penalty \citep{wang2020alignment}; the contrastive temperature follows a self-paced schedule adapted from \citet{zhang2024camel}. \textbf{Arm~S} (supervised) instead optimizes a paired margin objective on labeled metaphor/literal pairs sharing a target word, requiring $\Wmet{}(\text{fig}) - \Wmet{}(\text{lit}) \geq m$. Both arms apply the hierarchical regularizer $R_{\text{hier}}^{\text{met}} = \lVert \pi_{\text{cult}}(\zmet) - \zcult \rVert_2^2$, which keeps the metaphorical layer anchored to the cultural layer, and neither uses a classifier head: metaphoricity is read out geometrically (\S\ref{sec:wmet}). Loss terms, corpus details, and hyperparameters are given in Appendix~\ref{app:hyperparams}.

\subsection{Metaphorical Weight}
\label{sec:wmet}

CAMMAR estimates metaphoricity geometrically as the divergence between the lexical layer and the metaphorical layer mapped back into the lexical frame by the readout projector $g_{\text{m}\to\text{l}}$ (\S\ref{sec:arch}). For an input $x$,
\begin{equation}
  W_{\text{met}}(x) = 1 - \cos\!\left(z_{\text{lex}}(x),\; g_{\text{m}\to\text{l}}(z_{\text{met}}(x))\right),
\end{equation}
with both arguments in $\mathbb{R}^{768}$. A value near zero means the metaphorical layer stays aligned with its lexical anchor (literal); a value approaching one means the two have diverged (figurative). At the token level we apply the same readout per token $1 - \cos(z_{\text{lex}}(t), g_{\text{m}\to\text{l}}(z_{\text{met}}(t)))$  and mean-aggregate over the stem subword tokens of a target word to obtain its per-word weight $W_{\text{met}}(w\mid c)$, our primary evaluation signal. It is a \emph{measurement}, not a trained classifier: given the heads and the readout projector, it is computed at inference time with no example-level labels and no additional learned head.

\subsection{Evaluation Protocol}
\label{sec:eval}

We construct a span-annotated gold test set through an LLM-assisted, author-curated pipeline: an LLM proposes contextualized sentences for hand-curated seed target words, the author inspects each at generation time (discarding/regenerating any that misuse the target word, use dialect, or are ungrammatical), and each metaphorical record is paired with a literal companion using the same target word. Full protocol and prompts are in Appendix~\ref{app:goldset}.

Our evaluation is paired (Fig.~\ref{fig:evalpipeline}). Since the gold set is uniformly figurative, with negatives supplied by each record's literal companion, we score over the 184 matched pairs: the balanced ROC AUC, and the within-pair difference $\Delta\Wmet = W_{\text{met}}(w \mid c_{\text{met}}) - W_{\text{met}}(w \mid c_{\text{lit}})$, tested against zero with the one-sided Wilcoxon signed-rank test.

As a supplementary analysis, we probe each layer for its intended specialization with $\ell_2$-regularized logistic probes (Appendix~\ref{app:probing}).

% ============================================================================
\section{Experimental Setup}
\label{sec:setup}
% ============================================================================

\paragraph{Backbone.} CAMMAR is built on \texttt{U4RASD/NeoAraBERT\_MSA} (300M parameters), a NeoAraBERT-family encoder pretrained on Modern Standard Arabic. Subspace dimensionalities follow the nesting \(d_{\mathrm{lex}}{=}768 \supset d_{\mathrm{cult}}{=}384 \supset d_{\mathrm{met}}{=}192\) (\S\ref{sec:method}).

\paragraph{Training.} All phases use AdamW (weight decay 0.01), gradient clipping at norm 1.0, linear warmup with cosine decay, and mixed-precision training, with the per-layer similarity computations in the Phase-3 objectives kept in fp32 for numerical stability. \textbf{Phase 1} (lexical) trains the encoder and \Wlex{} with masked language modeling on a 100k-sentence Arabic Wikipedia subsample for 2 epochs at learning rate $5{\times}10^{-5}$, batch size 32. We train two Phase-1 variants that differ only in this phase: \emph{Model~A} uses MLM alone, and \emph{Model~B} adds the token-level root-consistency loss ($\alpha_{\mathrm{root}}{=}0.3$) with roots from the ALMA analyzer cached offline. \textbf{Phase 2} (cultural) freezes the encoder and \Wlex{} and trains \Wcult{} on the Anchor Context Cultural Association Bank (\S\ref{sec:phases}). \textbf{Phase 3} (figurative) freezes the encoder, \Wlex{}, and \Wcult{}, and trains \Wmet{} and the readout projector $g_{\text{m}\to\text{l}}$ for 15 epochs at learning rate $2{\times}10^{-5}$, batch size 32, temperature 0.07. We run two Phase-3 \emph{arms} from each Phase-1 model: \emph{Arm~U} optimizes a label-free token-level domain contrast (poetry vs.\ an arwiki literal control at a 0.3 ratio) with a uniformity term; \emph{Arm~S} optimizes a paired margin objective on labeled metaphor/literal pairs ($\Wmet{}(\text{fig}) - \Wmet{}(\text{lit}) \geq m$, $m{=}0.1$). The readout projector is re-initialized from a fixed seed at the start of each arm and is excluded from the hierarchical regularizer (\S\ref{sec:arch}). All runs use seed 42 on a single NVIDIA RTX 3090 (24GB).

\paragraph{Supervised-head reference.} As an upper-bound reference (Table~\ref{tab:ablation}, last row), we train a linear metaphor-detection head on \zmet{} at the target-word span on a 563-example supplementary set (50 seed words, disjoint by target word from the gold set), frozen backbone and heads; this is a trained classifier, not a geometric readout, and is reported only to bound what a supervised detector reaches on the same data.

\paragraph{Evaluation set.} The gold test set is scored as 184 matched pairs (\S\ref{sec:eval}); roots are resolved with the ALMA cache, gold target words are disjoint from all training data at the lemma and root level, and the set is used only at evaluation time.

\paragraph{Baselines.} We do not report comparisons against alternative architectures (vanilla MRL, SMEC-style nested training, or a directly fine-tuned supervised baseline); see Limitations for discussion.

% ============================================================================
\section{Results}
\label{sec:results}
% ============================================================================

We evaluate on the gold set (\S\ref{sec:setup}): 184 matched pairs sharing a target word, so the comparison holds word identity fixed and isolates figurative vs.\ literal \emph{usage}. \Wmet{} is read out at the target-word span on this held-out set. We test two questions in a $2\times2$ comparison: whether metaphoricity is legible in the geometry and under what supervision (Arm~U vs.\ Arm~S), and whether root grounding improves the readout (Model~A vs.\ Model~B), with every other ingredient held identical.

\begin{table*}[t]
  \centering
  \small
  \begin{tabular}{llcc}
    \toprule
    & & \multicolumn{2}{c}{Geometric \Wmet{} (gold)} \\
    \cmidrule(lr){3-4}
    Phase 1 & Phase 3 arm & AUC & paired $p$ \\
    \midrule
    A (MLM)      & U (unsupervised) & 0.479 & $0.87$ \\
    A (MLM)      & S (supervised)   & 0.795 & $1.1\times10^{-19}$ \\
    B (MLM+root) & U (unsupervised) & 0.415 & $1.0$ \\
    B (MLM+root) & S (supervised)   & \textbf{0.840} & $4.0\times10^{-24}$ \\
    \midrule
    \multicolumn{2}{l}{Supervised head (upper bound)} & 0.99 & $<10^{-30}$ \\
    \bottomrule
  \end{tabular}
  \caption{Geometric metaphor detection on the 184-pair gold set, as a $2\times2$ ablation over the Phase-1 objective (root grounding) and the Phase-3 arm (supervision of the geometry). AUC is the balanced ROC AUC over the paired fig/lit records; $p$ is the one-sided Wilcoxon signed-rank test on the within-pair difference $\Delta\Wmet{}$. The supervised classifier head (not a geometric readout) is shown as an upper-bound reference. All Phase-3 runs use 15 epochs.}
  \label{tab:ablation}
\end{table*}

\subsection{Metaphoricity is legible in the geometry, but only under supervision}
\label{sec:results-arms}

Table~\ref{tab:ablation} reveals a clear separation between the two arms. Arm~S detects metaphor well above chance: Model~A achieves AUC $0.795$ with $78.3\%$ concordance (median $\Delta\Wmet{} = +0.052$, $p = 1.1\times10^{-19}$), and Model~B reaches AUC $0.840$ with $82.6\%$ concordance (median $\Delta\Wmet{} = +0.062$, $p = 4.0\times10^{-24}$). By contrast, Arm~U remains at chance for both models (AUC $0.479$ and $0.415$; $p = 0.87$ and $1.0$). Because the paired design controls the target word, register, and context (e.g., \arb{أسد} used metaphorically versus literally), the Arm~S advantage cannot be explained by lexical or genre confounds.

This contrast supports the central claim of frame~(2): metaphoricity is \emph{legible} in the divergence between lexical and metaphorical layers (Fig.~\ref{fig:wmet-dist}), but it does not \emph{emerge} from label-free domain contrast. Instead, paired supervision must shape the geometry before the training-free readout can recover it, consistent with evidence that unsupervised contrastive objectives may solve their proxy task while weakly encoding the downstream property \citep{oord2018cpc, wang2020alignment}. We therefore characterize CAMMAR as a \emph{supervised-contrastive geometry with a training-free readout}. While a supervised classifier reaches AUC $0.99$ (Table~\ref{tab:ablation}), the geometric readout recovers a strong paired effect ($p < 10^{-19}$) without requiring a classifier at inference time.

\subsection{Root grounding gives a small, consistent gain}
\label{sec:results-root}

At a matched 15-epoch budget, the root-grounded Model~B outperforms the MLM-only Model~A in the supervised-geometry arm by a modest but consistent margin: $+0.045$ AUC ($0.840$ vs.\ $0.795$), $+4.3$ points of pair concordance ($82.6\%$ vs.\ $78.3\%$), a larger median effect ($+0.062$ vs.\ $+0.052$), and a smaller $p$-value ($4.0\times10^{-24}$ vs.\ $1.1\times10^{-19}$). The improvement is small in absolute AUC but moves every metric in the same direction, which is the pattern expected of a real effect rather than run-to-run noise. In the unsupervised arm, where the geometry carries no metaphor signal to begin with, root grounding makes no difference (both at chance).

\paragraph{The benefit is anchor quality, not visible root clustering.} A direct diagnostic complicates the naive reading of this result. Probing the lexical layer for root organization mean within-root versus across-root cosine separation, and nearest-neighbor root purity shows that Models~A and~B are near-identical and both only weakly root-organized (separation $+0.037$ vs.\ $+0.026$; NN purity $0.14$ vs.\ $0.11$, against a chance of $0.04$), with the projection in fact \emph{discarding} part of the root signal already present in the raw encoder ($+0.07$ separation). In other words, the root objective does not make \zlex{} visibly cluster by root any more than MLM alone does, yet it still improves downstream metaphor detection. We read this as evidence that root grounding improves the \emph{anchor quality} of \zlex{} its suitability as the reference frame against which \zmet{} is measured through a mechanism not captured by cosine clustering of the layer in isolation. This probe-versus-task divergence is a cautionary methodological point: a representation can be a better anchor for a downstream geometric readout without being more separable under direct probing.

\subsection{The cultural layer learns its associations but generalizes weakly}
\label{sec:results-cultural}

We evaluate the cultural layer as concept retrieval on a human-verified gold set of 169 anchor-context records. Although it fits its training associations almost perfectly (retrieval accuracy $\approx 1.0$), generalization to held-out anchors remains limited: top-1 retrieval reaches $0.059$ versus a $0.014$ chance rate ($4.2\times$ lift), increasing to $19\%$ at top-5. Allowing credit for \emph{any} valid concept among an anchor's multiple associations raises top-1 accuracy only to $0.095$. Thus, the cultural layer reliably memorizes training associations but does not yet generalize robustly to unseen anchors. We present it as a method and resource, leaving robust cultural generalization as future work, likely requiring broader per-concept coverage or partial encoder adaptation in Phase~2.

% ============================================================================
\section{Discussion}
\label{sec:discussion}
% ============================================================================

The decisive factor is not the readout but whether paired supervision shapes the geometry: label-free domain contrast solves its proxy objective yet transfers only at chance to metaphoricity, consistent with unsupervised contrastive learning where proxy objectives weakly encode the target property \citep{oord2018cpc, wang2020alignment}. We therefore characterize CAMMAR as a supervised-contrastive geometry with a training-free readout, with root grounding strengthening the lexical layer as a \emph{measurement anchor} rather than improving its standalone separability. We remain cautious in interpretation: the readout falls short of a supervised classifier (AUC $0.99$), and both the supervision pairs and gold set comprise LLM-generated \arb{فصحى} sentences from the same pipeline, so shared artifacts cannot be excluded. Even so, for Arabic, where large sentence-level metaphor corpora are scarce, metaphoricity can be measured directly from inter-layer geometry using paired supervision over only dozens of words, without a classifier.

% ============================================================================
\section{Conclusion}
\label{sec:conclusion}
% ============================================================================

Figurative meaning in Arabic is often flattened by representations that conflate lexical, cultural, and metaphorical information. We introduced CAMMAR, which separates meaning into three nested embedding subspaces trained through a sequential semantic curriculum with controlled freezing, rather than the identical-supervision scheme of standard Matryoshka representation learning. The design operationalizes \citeposs{jurjanidalail} theory of \emph{na\d{z}m} and yields a geometric, training-free metaphoricity readout \Wmet{}. On a gold set of 184 word-matched figurative/literal pairs, \Wmet{} detects metaphor well above chance under supervised-contrastive training (AUC up to $0.840$, $p<10^{-23}$) but remains at chance under unsupervised domain contrast. Root grounding provides a small but consistent gain ($+0.045$ AUC) visible only in the geometric readout, not direct probing. We will release the model, gold set, and construction pipeline upon acceptance.

\paragraph{Future directions.} The gold set could be expanded from classical rhetorical sources (\emph{Bal\=agha} textbooks and the poetic-commentary \emph{shur\=u\d{h}} tradition); the cultural layer could be deepened on the substrates Arabic figuration draws on and on multiple parallel cultural reservoirs, enabling models of \emph{cross-cultural} divergence. The concept inventory is, by design, the node set of a future Cultural Knowledge Graph: turning the Bank's distributional associations into explicit typed relations (\emph{symbolizes}, \emph{embodies}, \emph{contrasts-with}) would add multi-hop cultural inference.

% ============================================================================
% Limitations is a REQUIRED unnumbered section for *ACL submissions.
% Per the template: use \section* (not \section) so the section is
% unnumbered, and place it AFTER the Conclusion but BEFORE the bibliography.
% ============================================================================
\section*{Limitations}

\textbf{Scale and baselines.} The gold set yields 184 figurative/literal minimal pairs sufficient for the paired tests we report, but smaller than English benchmarks such as VUA-MIPVU, and we prioritize span-level annotation quality over scale. We also do not compare against alternative architectures (a vanilla MRL ablation without the sequential curriculum, SMEC-style nested training, or a directly fine-tuned supervised baseline). Our results therefore establish that CAMMAR's nested geometry carries a real, supervision-legible metaphoricity signal, but not that it exceeds what a simpler architecture would achieve on the same data; both comparisons are direct next steps.

\textbf{Supervision proxy and a one-dimensional readout.} Arm~U derives its signal from corpus identity (figurative-leaning poetry vs.\ a literal-leaning Wikipedia control), which is only a proxy for figurativity: poetry contains denotative description and encyclopedic prose contains conventionalized figuration, so the unsupervised arm conflates figurative abstraction with stylistic register consistent with its chance-level result. Separately, \Wmet{} is a single scalar that captures \emph{how figurative} a word is but does not distinguish the classical metaphor categories (\emph{isti'\=ara}, \emph{kin\=aya}); recovering those would require additional structured supervision.

\textbf{Coverage and backbone.} All training corpora and the gold set are restricted to Modern Standard and Classical Arabic; we do not evaluate dialectal varieties, whose figuration draws on different source domains. All results use a single backbone, NeoAraBERT-MSA \citep{abou-chakra-etal-2026-neoarabert}; the architecture makes no backbone-specific assumptions, but transfer across encoders is unverified. The cultural layer is evaluated as concept retrieval on a disjoint, human-verified gold set (\S\ref{sec:results}); it learns its training associations but generalizes only weakly to held-out anchors (top-1 $4.2\times$ chance), and is not yet used as direct cultural-metaphor supervision. Strengthening this generalization through broader per-concept coverage or partial encoder adaptation in Phase~2 is left to future work.

\textbf{Annotation subjectivity.} The gold set was built through an LLM-assisted pipeline with inline curation by a native Arabic speaker: every generated sentence was inspected at generation time and unsuitable ones discarded and regenerated. Because verification was inline rather than an independent post-hoc pass, we report no separate acceptance rate or inter-annotator agreement; the resource reflects single-curator judgment, and independent multi-annotator validation is left to future work.

\section*{Ethics Statement}
 
\textbf{Data sources and licensing.} Our training corpora consist of Arabic Wikipedia and publicly available classical Arabic poetry and heritage texts, used for non-commercial research. The cultural association data and the span-annotated metaphor set were generated with large language models and curated by the authors; we do not redistribute any copyrighted source text, and released artifacts will contain only model-generated sentences and our own annotations.
 
\textbf{Use of large language models.} The metaphor gold set and the cultural association bank were produced through LLM-assisted generation (with Anthropic, OpenAI, or Google backends) followed by human curation. LLM-generated text can carry the biases and factual errors of its source models; we mitigate this through inline native-speaker review.
 
\textbf{Cultural representation.} CAMMAR encodes conventional cultural associations (e.g.\ figures and symbols of generosity, courage, or treachery) drawn predominantly from a Modern Standard and Classical Arabic literary tradition. This is a partial and historically situated view of ``Arabic culture'': it underrepresents dialectal, regional, and contemporary figuration, and the fixed concept inventory reflects choices made by the authors. The cultural layer should not be read as a complete or normative model of cultural meaning, and we caution against deploying it in settings where such associations could reinforce stereotypes about groups or individuals.
 
\textbf{Intended use and risks.} CAMMAR is a research framework for studying how figurative meaning is represented; it is not a deployed system for high-stakes decisions. Its metaphoricity readout is a continuous geometric measurement, not a verified classifier, and the cultural layer generalizes only weakly to unseen anchors (\S\ref{sec:results-cultural}); both should be used with appropriate caution and not as ground truth about whether a given utterance is figurative or what it culturally connotes.
 
\textbf{Reproducibility and release.} To support reproducibility, we report seeds, hyperparameters, and corpus statistics (\S\ref{sec:setup}, Appendix~\ref{app:hyperparams}) and will release the datasets, the cultural concept inventory, and code upon acceptance, with documentation of the generation and curation pipeline (Appendix~\ref{app:goldset}). All experiments were run on a single consumer GPU, keeping the computational and environmental footprint modest.

% ============================================================================
% Acknowledgments is an optional unnumbered section, also placed before the
% bibliography. In review mode (\usepackage[review]{acl}) you should usually
% leave this commented out or use a non-identifying placeholder, since named
% acknowledgments would break anonymity. Uncomment and fill for the final
% camera-ready version.
% ============================================================================
% \section*{Acknowledgments}
% \TODO{Acknowledgments for camera-ready version only. Funding sources,
% institutional support, anonymous reviewers, advisors, computing resources.}

% ============================================================================
%  Bibliography
% ============================================================================
\bibliography{refs}

% ============================================================================
%  Appendix
% ============================================================================
\appendix

\section{Gold Set Construction Protocol}
\label{app:goldset}

The span-annotated metaphor gold set is built in four stages (G1-G4) through an LLM-assisted pipeline with inline single-curator verification. The same infrastructure (a provenance-tracked record store, a pluggable LLM call supporting manual paste-back and automated backends, and diacritic-tolerant span location) is reused for the Cultural Association Bank (\S\ref{sec:phases}).

\paragraph{G1: Seed collection.} We curate a list of seed target words, each a single content word with a well-attested metaphorical sense in Modern Standard or Classical Arabic. Every seed carries its target word, a literal gloss, a metaphorical gloss (with an optional Arabic rhetorical note), and its source and target conceptual domains (e.g.\ \arb{أسد}: source \emph{animal}, target \emph{human courage}). Seeds are drawn from standard rhetorical inventories and lexicographic sources; classical-poetry candidate verses are additionally collected as raw material for verse-based target extraction.

\paragraph{G2: LLM-assisted proposal generation.} For each seed, an LLM is prompted to produce several contextualized sentences in which the target word is used in its metaphorical sense. Generation runs in either a manual mode (the rendered prompt is shown and the model's reply pasted back) or an automated mode (Anthropic, OpenAI, or Google backends), with identical downstream parsing. Three prompt templates are used: a target-word contextualization prompt that elicits metaphorical sentences for a seed; a paired-literal-companion prompt that, given a metaphorical sentence, elicits a literal-sense sentence using the \emph{same} target word; and a verse target-extraction prompt that identifies a metaphorically-used target word and its span within a classical verse. Generation is resumable, tracking completed seeds by provenance.

\paragraph{G3: Inline curation.} Every proposed sentence is inspected at generation time by a native-speaker curator. A sentence is discarded (and regenerated) if it misuses or omits the target word, lapses into dialect, is ungrammatical, or fails to realize the intended figurative sense. Accepted sentences have their target span located automatically (with a diacritic-relaxed fallback) and validated against the surface form. Dialectal sentences are filtered by a heuristic marker list as a second guard.

\paragraph{G4: Finalization.} Each metaphorical record is paired with a literal companion sharing the same target word, linked by record identifier, so that evaluation can control for word identity. Records carry the sentence text, the target span (character offsets, surface form, and analyzer-derived root), an expected label, pairing links, and full generation provenance. The final set comprises 1{,}098 sentences 914 metaphorical and 184 literal over 184 target words, scored as 184 word-matched figurative/literal pairs (\S\ref{sec:eval}). Verification is by a single curator inline during generation; multi-annotator agreement is left to future work. Target words in the gold set are disjoint from all training corpora at the lemma and root level.

\section{Hyperparameter Details}
\label{app:hyperparams}

\begin{table}[h]
\centering
\small
\begin{tabular}{lll}
\toprule
Phase & Key settings \\
\midrule
1 (lexical) & MLM; 2 ep; lr $5{\times}10^{-5}$; bs 32; \\
            & Model~B adds root SupCon, $\alpha_{\mathrm{root}}{=}0.3$ \\
2 (cultural) & frozen enc.\ + \Wlex{}; anchor$\to$concept \\
             & InfoNCE + SupCon + uniformity; $\tau{=}0.07$ \\
3 (figurative) & frozen enc./\Wlex{}/\Wcult{}; 15 ep; \\
               & lr $2{\times}10^{-5}$; bs 32; $\tau{=}0.07$; \\
               & Arm~S margin $m{=}0.1$; Arm~U $\lambda_{\mathrm{unif}}{=}0.1$ \\
\midrule
all & AdamW, wd 0.01, clip 1.0, warmup+cosine, \\
    & fp16 (fp32 similarities), seed 42, RTX 3090 \\
\bottomrule
\end{tabular}
\caption{Training hyperparameters per phase.}
\end{table}

\section{Layer-Specialization Probing}
\label{app:probing}
To test whether the three layers carry their intended content, we probe each layer $z_k$ for three tasks: lexical (predict the target word's root), cultural (predict the concept from the inventory, \S\ref{sec:phases}), and figurative (predict metaphoric use). We standardize embeddings, fit an $\ell_2$-regularized logistic regression ($C{=}1$), and report 5-fold stratified accuracy. The per-layer specialization is
\begin{equation}
  \Delta_k = \mathrm{acc}(z_k, T_k^\star) - \tfrac{1}{2}\!\sum_{T \neq T_k^\star}\! \mathrm{acc}(z_k, T),
\end{equation}
with $T_k^\star$ the task aligned to $z_k$, and the global index $\Delta = \tfrac{1}{3}\sum_k \Delta_k$; positive $\Delta$ means each layer leads on its own task.

\section{Supplementary Figures}

\begin{figure*}[t]
  \centering
  \includegraphics[width=\textwidth]{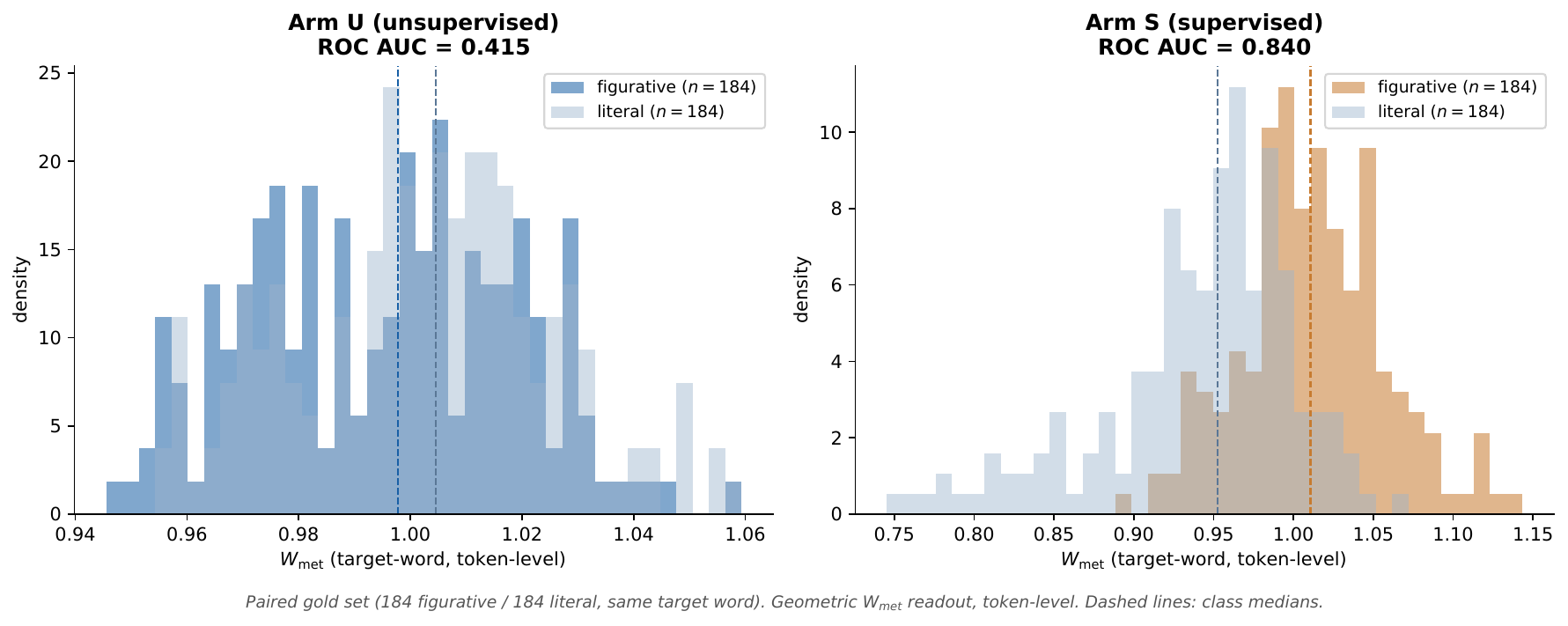}
  \caption{Distribution of the geometric metaphoricity readout \Wmet{} over the 184 matched figurative/literal pairs (token-level, target-word span). \emph{Left:} Arm~U (unsupervised domain contrast) places figurative and literal uses in overlapping distributions (AUC near chance). \emph{Right:} Arm~S (supervised paired margin) separates them (AUC $0.840$ for Model~B). Dashed lines mark class medians. The contrast visualizes the central finding: metaphoricity is legible in the inter-layer geometry only when paired supervision shapes it.}
  \label{fig:wmet-dist}
\end{figure*}

\label{app:figures}
\begin{figure}[t]
  \centering
  \includegraphics[width=\columnwidth]{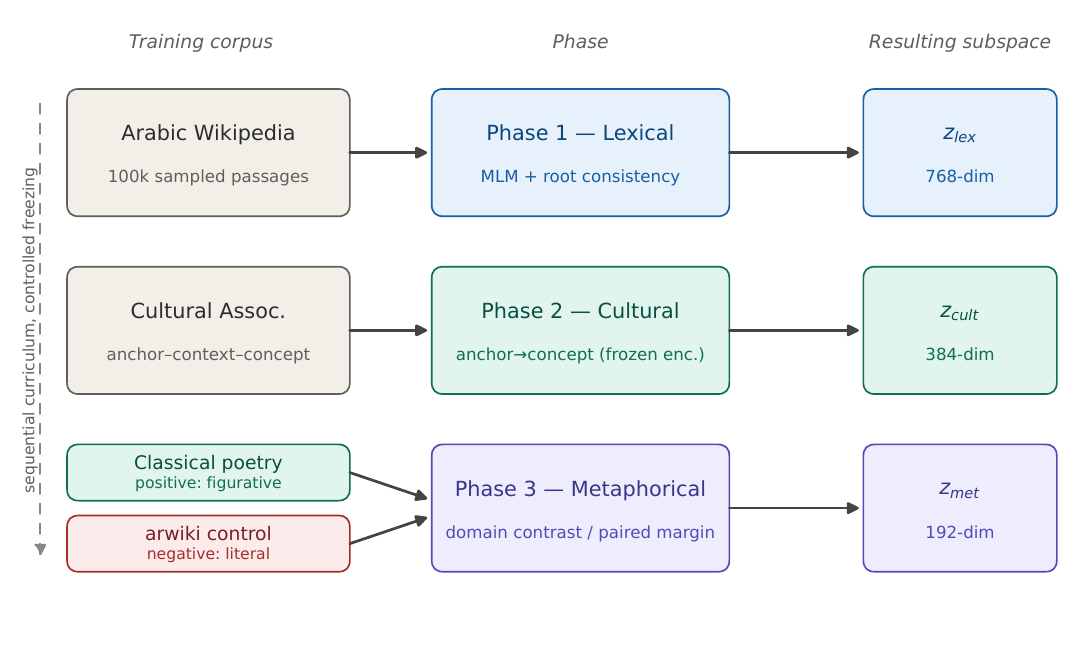}
  \caption{Training data flow across CAMMAR's three phases. Phase 1 uses Arabic Wikipedia for masked language modeling, with an optional token-level root-consistency objective (Model~B) using analyzer-derived roots. Phase 2 builds the cultural layer from the Anchor Context Cultural Association Bank, training anchor-to-concept retrieval with a frozen encoder. Phase 3 trains the metaphorical layer in one of two arms: an unsupervised domain contrast (poetry vs.\ an arwiki literal control) or a supervised paired margin objective on labeled metaphor/literal pairs. The dashed arrow indicates the sequential curriculum.}
  \label{fig:dataflow}
\end{figure}

\begin{figure}[t]
  \centering
  \includegraphics[width=\columnwidth]{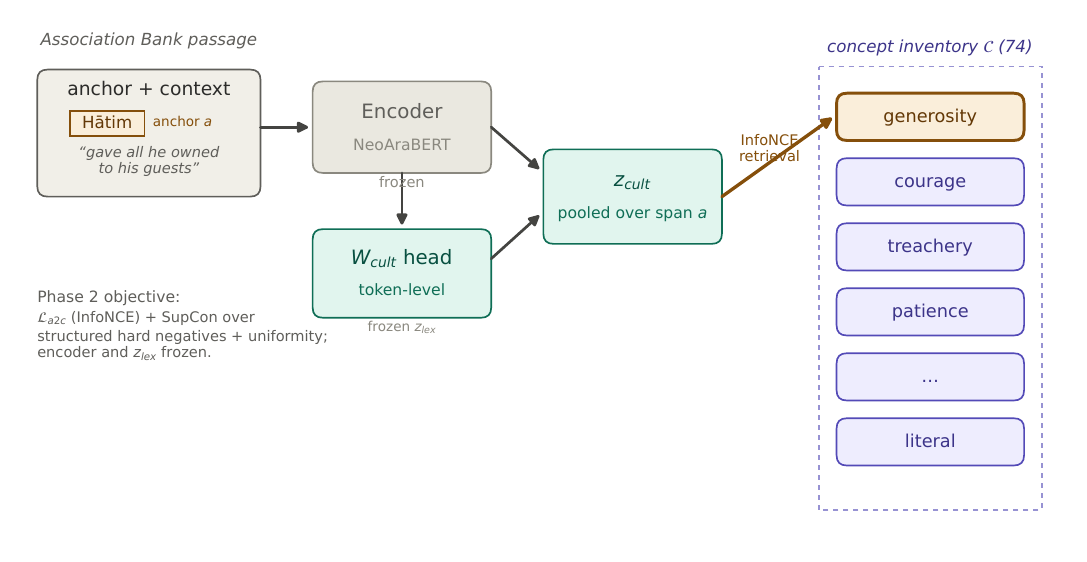}
  \caption{The cultural layer (Phase~2, \S\ref{sec:phases}). An Association Bank passage pairs a culturally loaded anchor entity (e.g.\ \arb{حاتم}) with the surrounding context that activates its conventional association. The frozen encoder and \Wcult{} head produce a token-level cultural representation pooled over the anchor span ($z_{\mathrm{cult}}$), which is matched by InfoNCE retrieval against a fixed inventory of 74 concept prototypes (here, \textsc{generosity}). The full Phase-2 objective additionally includes a supervised-contrastive term over structured hard negatives and a uniformity penalty; the encoder and \zlex{} are frozen so the literal layer is preserved. The figure shows the retrieval path only.}
  \label{fig:cultural}
\end{figure}

\begin{figure}[t]
  \centering
  \includegraphics[width=\columnwidth]{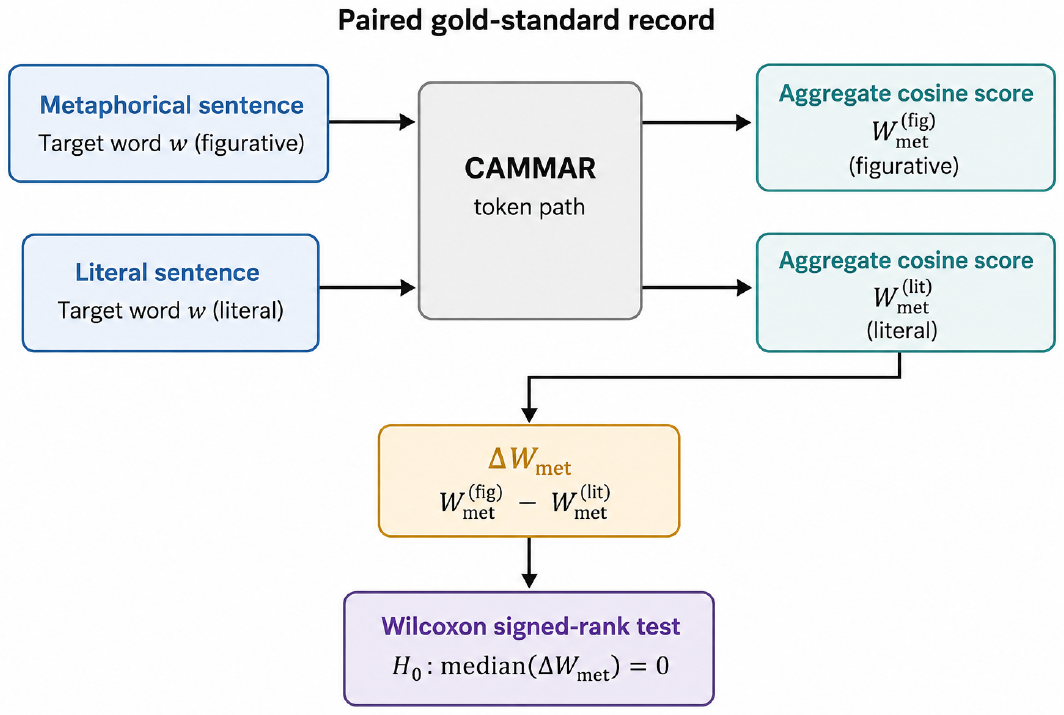}
  \caption{CAMMAR's evaluation pipeline at the token level. Each paired gold record consists of a metaphorical sentence and a literal companion sharing the same target word. Both sentences are encoded through CAMMAR's token path, after which the per-target-word metaphorical weight \Wmet{} is computed by aggregating per-token cosine distances over the target span. The paired difference $\Delta\Wmet$ should be reliably positive across the paired subset, tested with the Wilcoxon signed-rank test.}
  \label{fig:evalpipeline}
\end{figure}

\end{document}